\documentclass[runningheads]{llncs}

\usepackage[preprint,year=2024,ID=6581]{eccv}
\usepackage{wrapfig}

\usepackage{eccvabbrv}

\usepackage{graphicx}
\usepackage{booktabs}

\usepackage{caption}
\usepackage{subcaption} 
\usepackage{makecell}
\usepackage{threeparttable}

\usepackage[accsupp]{axessibility}  

\usepackage{multirow}
\usepackage{colortbl}

\definecolor{darkred}{rgb}{0.7, 0.0, 0.0}
\definecolor{darkred2}{rgb}{0.5, 0.0, 0.0}
\definecolor{darkred3}{rgb}{0.9, 0.0, 0.0}
\definecolor{darkgreen}{rgb}{0.0, 0.42, 0.24}
\definecolor{darkblue}{rgb}{0.10, 0.17, 0.8}
\definecolor{Gray}{gray}{0.93}

\usepackage[pagebackref,breaklinks,colorlinks,citecolor=eccvblue]{hyperref}

\usepackage{orcidlink}

\begin{document}

\title{An End-to-End Optimized Lensless System \\ for Privacy-Preserving Face Verification} 

\titlerunning{LenslessFace}

\author{
  Xin Cai$^{1,2}$, Hailong Zhang$^3$, Chenchen Wang$^{1,3}$, Wentao Liu$^{1,4}$, Jinwei Gu$^{2}$, Tianfan Xue$^{2}$ 
}
\authorrunning{X.Cai et al.}

\institute{  $^1$Shanghai AI Laboratory, $^2$The Chinese University of Hong Kong, $^3$Tsinghua University, $^4$SenseTime\\
  {\tt\small \{cx023, tfxue\}@ie.cuhk.edu.hk, jwgu@cuhk.edu.hk,} \\ 
  {\tt\small \{zhanghl21,wcc20\}@mails.tsinghua.edu.cn, liuwentao@sensetime.com}}

\maketitle
\vspace{-2em}
\begin{abstract}
Lensless cameras, innovatively replacing traditional lenses for ultra-thin, flat optics, encode light directly onto sensors, producing images that are not immediately recognizable. This compact, lightweight, and cost-effective imaging solution offers inherent privacy advantages, making it attractive for privacy-sensitive applications like face verification.
Typical lensless face verification adopts a two-stage process of reconstruction followed by verification, incurring privacy risks from reconstructed faces and high computational costs.
This paper presents an end-to-end optimization approach for privacy-preserving face verification directly on encoded lensless captures, 
ensuring that the entire software pipeline remains encoded with no visible faces as intermediate results. To achieve this, we propose several techniques to address unique challenges from the lensless setup which precludes traditional face detection and alignment. Specifically, we propose a face center alignment scheme, an augmentation curriculum to build robustness against variations, and a knowledge distillation method to smooth optimization and enhance performance. Evaluations under both simulation and real environment demonstrate our method outperforms two-stage lensless verification while enhancing privacy and efficiency.  Project website: \url{lenslessface.github.io}.
\end{abstract}
   
\vspace{-2em}
\section{Introduction}
\label{sec:intro}
\vspace{-1em}

\begin{figure}[t]
\centering
\includegraphics[width=\linewidth]{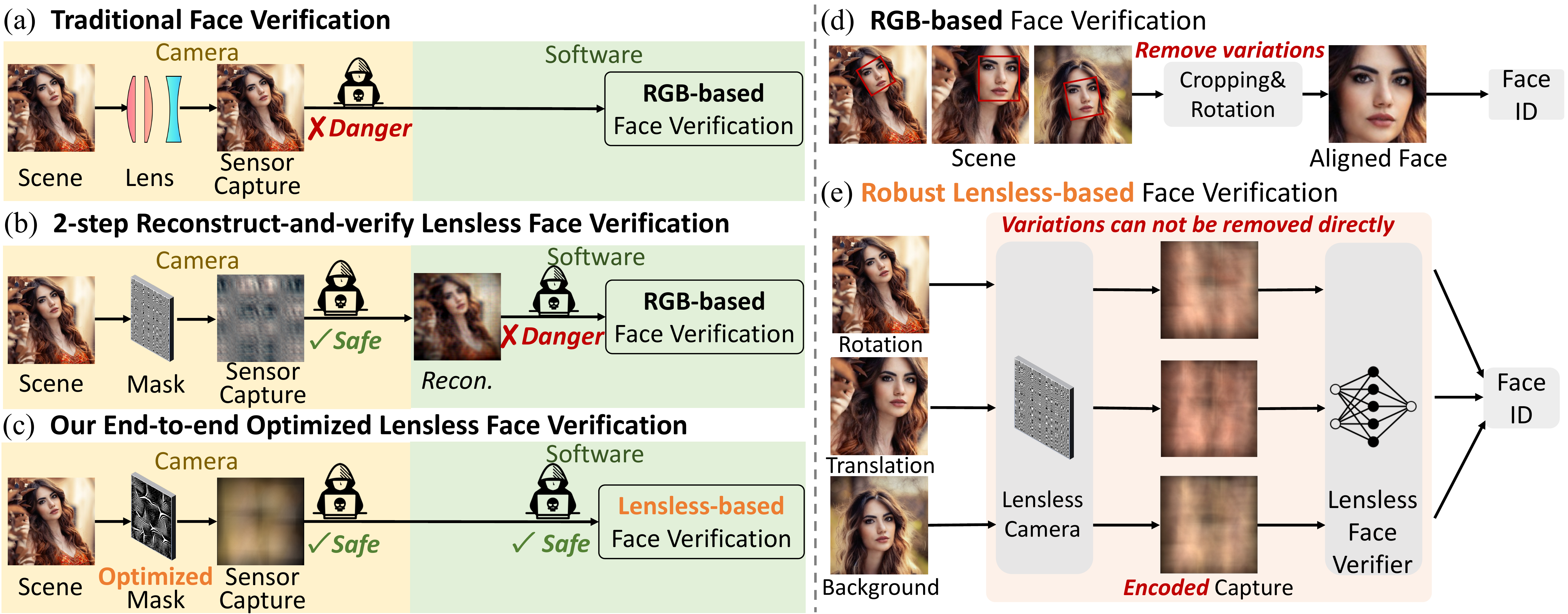}
\vspace{-1.5em}
\caption{
\textbf{Left: Comparison of different face verification approaches}: lens-based (top) v.s. two-step lensless (middle) v.s. our optimized one-step lensless method. Our approach enhances privacy against software attacks as well as maintains robust face verification performance.
\textbf{Right:
Challenges for robust lensless-based face verification as compared to RGB-based methods.}
Our robust lensless-based face verification is designed to accurately distinguish identities with encrypted captures, demonstrating resilience to variations in background, rotation, shift, and scales.
}
\vspace{-2.em}
\label{fig:teaser}
\end{figure}

With recent advances in optical engineering and computational photography, lensless imaging systems have emerged as a new, ultra-compact, lightweight, and cost-effective imaging system. Unlike traditional cameras that use large, heavy-weight focusing lenses, lensless cameras capture light using a simple and flat optic, such as an amplitude \cite{asif2016flatcam,bezzam2022privacy} or phase mask \cite{boominathan2020phlatcam,antipa2018diffusercam,kuo2020chip}. These optics transform incoming light into highly multiplexed patterns. Consequently, images captured by lensless cameras are not immediately recognizable and require detailed knowledge of the system's mask pattern to reconstruct the original captured scenes.
The low cost, small size, lightweight, and potential for increased privacy make lensless imaging systems a promising option for use in wearable devices, augmented and virtual reality, and biometric identification.

Specifically, lensless cameras offer a promising hardware solution for privacy-preserving imaging. Taking face recognition as an example, traditional recognition systems capture a clear face image using a convergence lens camera and recognize facial identity through a verification algorithm, as shown in~\cref{fig:teaser}(a). Such a verification system may be vulnerable to the potential of a network attack, which may hijack camera data. On the other hand, lensless cameras only capture unrecognizable images and thus protect facial images from security attacks.

A typical lensless verification system uses a \textit{reconstruct-and-verify} approach. This approach first restores the original image from the encoded one and then subsequently applies conventional verification algorithms. However, this method has its weaknesses. First, while lensless imaging can prevent hackers from hijacking camera data during transmission, it remains vulnerable to software attackers who can directly access reconstructed images, as illustrated in Figure \ref{fig:teaser}(b). Second, the recognition accuracy is much lower than that of traditional RGB face verification systems, both because the reconstructed images are often blurry and the mask patterns designed for reconstruction are sub-optimal for face verification. Lastly, the additional reconstruction is not mandatory and may increase computational costs, making it a less desirable solution for edge computing.

Therefore, to improve verification accuracy while preserving privacy better, we propose a novel reconstruction-free method with lensless cameras for facial verification tasks. Unlike the reconstruct-and-verify process, our approach is a one-step solution that directly verifies faces on the encoded sensor images, without reconstructing facial images as intermediate results in the pipeline, as illustrated in ~Fig.\ref{fig:teaser}(c). This design has three advantages. First, the entire system is invulnerable to both camera data hijacking and software attacks, and it also reduces computational cost by skipping the reconstruction step. Second, the mask is also optimized to improve verification accuracy, capturing key facial features for verification.
Therefore, it outperforms hand-crafted masks that are only designed for high-quality reconstruction.
In summary, our approach is an end-to-end optical-electronic neural network for face verification, offering enhanced security, cost efficiency, and improved verification accuracy.

Moreover, a simple end-to-end learning strategy for the proposed optical-electronic verification may result in poor accuracy (70\% on LFW \cite{LFWTech}). To deal with face rotation, translation, or background changes, traditional face verification pipelines need to first align input images to a canonical pose using simple transformations and cropping, as shown in  ~\cref{fig:teaser}(d). However, in a lensless setup, it only captures an encoded image that does not keep the original spatial structure, making traditional alignment impossible, as shown in~\cref{fig:teaser}(e).

In response to these challenges, we propose a novel robust scheme to make end-to-end optimization with various face images practicable (from 70\% to 95\% on LFW). First, we design a new face-center alignment strategy, aligning the face center with the center of lensless images, enhancing system resilience against face translation.
Second, to make our model robust to variations including rotation, scale, and background changes, we adopt an augmentation curriculum of variations. This curriculum gradually increases the complexities of variations, allowing the model to initially focus on learning face verification and then progressively build robustness against these variations. Third, to circumvent the local minimum during training, we proposed a novel cross-modality knowledge distillation method.  This technique transfers verification capabilities from an RGB face verification model to our lensless verification model, smoothing the loss function landscape and aligning face verification performance to the RGB model. It is noted that the three strategies function as a holistic system, with each process being essential for achieving satisfactory optimization results.

In summary, our contributions are as follows:
1) We introduce an end-to-end optimization-based face verification system using lensless cameras. The proposed method brings improvements in privacy, accuracy, and speed when compared to conventional two-stage lensless face verification techniques.
2) We propose a novel training scheme to make the end-to-end verification network trainable. The training scheme includes a face center alignment tailored for lensless setup, a new cross-modality knowledge distillation to avoid local minima, and an augmentation curriculum that improves the robustness of our model under different variations. 
3). We evaluate the efficacy of our method in both simulated and real environments. Experiments show that our method outperforms the two-step solutions, moving closer to the accuracy of traditional RGB-based approaches while inheriting hardware-level privacy protections from the lensless design. 

\vspace{-1em}
\section{Related Work}
\vspace{-0.5em}
\subsection{Lensless imaging}
\vspace{-0.2em}
Lensless imaging \cite{boominathan2022recent} has become an emerging topic in visible light computational imaging due to its potential for miniaturizing camera systems. Unlike traditional cameras, lensless imaging systems use a mask element to modulate an incoming scene instead of a focusing lens. This technique can employ a variety of masks, including phase gratings \cite{stork2013lensless}, random diffusers \cite{antipa2018diffusercam,kuo2020chip}, hand-crafted designed phase masks \cite{chi2011optical,boominathan2020phlatcam}, amplitude masks \cite{asif2016flatcam,adams2017single,shimano2018lensless}, compressive samplers \cite{huang2013lensless,satat2017lensless}, and spatial light modulators \cite{zomet2006lensless,yuan2018parallel,hua2020sweepcam}. Replacing the lens with masks results in multiplexed sensor captures that lack visual resemblance to the scene. This provides inherent privacy protection as the original scene is non-visible.

Transforming lensless images into identifiable images usually involves the use of computational reconstruction algorithms, which can be broadly categorized into optimization-based methods and data-driven methods. Optimization based approaches model the reconstruction as a regularized least squares problem, with prominent examples such as sparsity in the spatial domain \cite{asif2016flatcam,boominathan2020phlatcam} and frequency domain \cite{reddy2011p2c2}. Conversely, data-driven methods \cite{rick2017one,khan2020flatnet,pan2022image} use deep neural networks to learn a mapping from the sensor captures to the reconstructed images.

\vspace{-1.5em}
\subsection{Perception with Lensless Camera}
\vspace{-0.7em}
Due to the inherent hardware-level privacy protection feature, lensless cameras have been widely utilized to perform perception tasks \cite{wang2019privacy,pan2021lensless,pan2021incoherent,ishii2020privacy,nguyen2019deep,henry2022aerial}, especially those associated with humans and faces, like face attribute detection \cite{bezzam2022privacy}, face recognition \cite{taheri2015cancelable,wu2022lensless,tan2018face,shi2022loen,henry2023privacy}, and action recognition \cite{wang2019privacy}. Some \cite{tan2018face,ishii2020privacy} adopt a two-stage reconstruction and perception pipeline, which could be suboptimal in terms of accuracy, privacy, and efficiency. Others \cite{pan2021lensless,pan2021incoherent,henry2023privacy} try to perform the perception tasks directly on encoded sensor images. However, due to the lack of mask optimization in the process, their performance remains inferior compared to traditional RGB-based methods. More recently, for enhanced privacy or performance, some work \cite{shi2022loen,bezzam2022privacy} have performed end-to-end optimization of lensless masks and neural networks for downstream perception tasks.

Our work is similar to \cite{shi2022loen,bezzam2022privacy} in performing joint optical-electric neural network optimization. However, Bezzam et al. \cite{bezzam2022privacy} optimize for privacy by downsampling encoded images, making the identities hard to distinguish. Shi et al. \cite{shi2022loen} only deal with aligned faces, and their face verification performance lags behind RGB-based methods. In contrast to these approaches, we develop a method for lensless-based face verification to withstand various variances, making it more suitable for real-world applications.

\vspace{-1.5em}
\section{End-to-end Lensless Optics Formation}
\vspace{-0.8em}
Deep optics \cite{sitzmann2018end,wetzstein2020inference}, or end-to-end methods for optimizing optical components, seek to co-design the optics and corresponding computational algorithm to maximize performance for a specific application. Based on this concept, we present an end-to-end-differentiable pipeline that supports joint optimization of the lensless camera mask and post-capturing processing network.

\textbf{Imaging formulation model and simulation pipeline.}
In our setup, we place an image sensor with a lensless mask at a known distance from the scene to be captured. Based on the scalar diffraction theory ~\cite{goodman2005introduction}, we have such an assumption: for a given wavelength $\lambda$, image formation between two parallel planes acts as a linear shift-invariant system. 
For incoherent light of a distant scene, the theory enables modeling the image formation $S_\lambda(x,y)$ on the sensor as a convolution:
\vspace{-.5em}
\begin{equation}
S_\lambda(x,y) = (l_\lambda * p_\lambda)(x,y),
\label{eq:image_form}
\vspace{-0.5em}
\end{equation}
where $*$ is the convolution operator, $l_\lambda(x,y)$ represents the scene's spatial intensity distribution. $p_\lambda(x,y)$ denotes the point spread function (PSF).
In our simulation, we use RGB images as the scene intensity distribution and simulate PSF based on the mask pattern. 
More details are discussed in the supplementary. 

Based on this model, the lensless sensor capture $S$ is formulated as a linear function $S= g(I, M)$ of the RGB image $I$ and the mask pattern $M$. This linear shift-invariant assumption is based on the scene-to-mask distance obeying the paraxial criterion~\cite{sitzmann2018end}. The majority of standard face verification scenarios meet this criterion, thereby validating the assumption for such applications.

\textbf{Trainable mask to PSF.} 
We formulate our lensless mask $M$ with learnable parameters for transparency $w(x, y)$ at varying positions $(x, y)$. The modulation for the amplitude $A_m(x, y)$  attributed to the incident wave is expressed as: $A_m(x, y) = w(x, y).$
We can optimize the mask parameters $\theta_M = \{w(x,y)\}$ to obtain the desired light modulations and PSF, hence enhancing the results of subsequent tasks. Note that in amplitude masks, the PSF $p_\lambda(x,y)$ remains consistent across any wavelength $\lambda$ and we can drop the wavelength parameter.

\begin{figure*}[t]
\includegraphics[width=\linewidth]{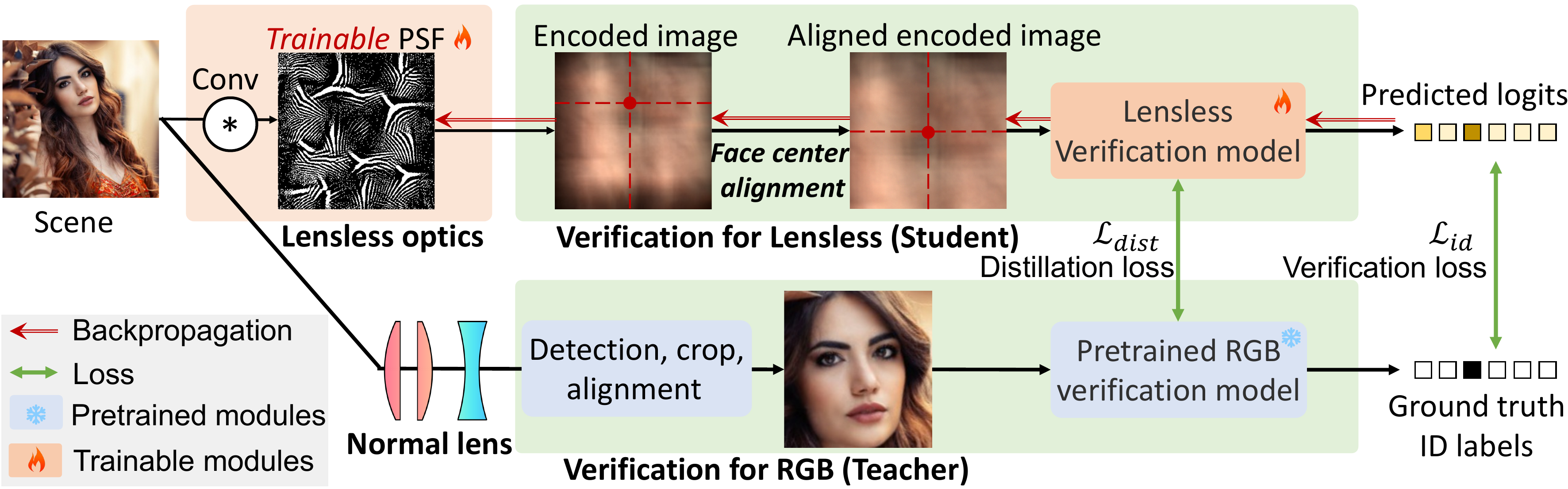}
\vspace{-2.0em}
\caption{
\textbf{Overview of our end-to-end optimization pipeline}. The top pathway simulates lensless imaging using a learnable mask to encode a facial scene into a sensor capture. The capture after face-center alignment is fed into the lensless student model. In the bottom pathway, the same scene is aligned by preprocessing, and then processed by a trained RGB teacher model. The teacher's output features and ID labels supervise the training of both the lensless student model and the learnable mask.
}
\label{fig:pipeline}
\end{figure*}

\textbf{End-to-end training objective.}
Leveraging a differentiable image formulation, we can simultaneously optimize the optical mask and electronic model for desired tasks. Unlike previous methods that only optimized the post-capture processing model, our end-to-end joint training also finds an optimal mask that extracts useful features for the target tasks. These features are stored in encoded images and are processed by the subsequent model, as illustrated in \cref{fig:pipeline}.

Given a training dataset $\{I_i,t_i\}_{i=1}^N$ with input scene image $I_i$ and the target label $t_i$, the optimization objective is:
\vspace{-.7em}
\begin{equation}\label{eq:target}
\min_{\theta,\theta_M}\sum_{i=1}^N\mathcal{L}(t_i,\mathcal{F}_\theta(g(I_i, M))) - \alpha\sum_{x,y}w(x,y),
\vspace{-.7em}
\end{equation}
where the first term $\mathcal{L}$ is for the loss function of the target task. For face verification, $\mathcal{L}$ is the ArcFace loss function~\cite{deng2019arcface}.
$\mathcal{F}_\theta$ is the downstream task model, like the face verification model, with parameter $\theta$.
$g(\cdot, \cdot)$ is the differentiable rendering process described above
The second term penalizes small masks, to ensure the learned mask's transparent regions are large enough to yield unidentifiable images for privacy-preserving. The $\alpha$ is a factor to balance the losses and in our experiments, we set  $\alpha=0.01$.

\vspace{-1.em}
\section{Lensless Face Verification Pipeline}

Our goal is to make the optical mask and face verification model optimal for face verification through an end-to-end training process. To increase efficiency and safeguard privacy, our pipeline skips the reconstruction process and directly executes face verification on encoded captures. Following the previous setups in \cite{tan2018face}, we assume that only one face appears in the captured scene.

One challenge of this pipeline is that the encoded captures it produces are not inherently suitable for face recognition. As shown in \cref{fig:teaser}(d), in traditional RGB face verification, the input images are first cropped and aligned to a canonical space before being sent to the verification model, to reduce the impact of rotation, translation, and cluttered backgrounds. However, this process cannot be applied to a lensless camera, as the encoded captures encapsulate all scene information, making direct cropping or alignment infeasible. A simple way to ensure robustness to different variations is using a training set that includes these variations or augmenting existing ones with such variations. Nonetheless, lensless captures lack the spatial structures vital for face recognition, which complicates the use of simple end-to-end optimization methods. Consequently, these methods alone have not been successful in establishing robust lensless face verification.

To train a robust face verification that can handle translation, rotation, and background changes, we propose a holistic end-to-end optimization framework comprising three critical components: 1) A novel physics-based alignment algorithm tailored for lensless captures, designed to counteract facial translations before processing by the face verification model; 2) A new augmentation curriculum to increase the robustness of the face verification model; 3) A cross-modality distillation method to ensure facial feature alignment and effective optimization.

\begin{figure}[t]
  \centering
  \begin{minipage}[b]{0.47\linewidth}
    \includegraphics[width=\textwidth]{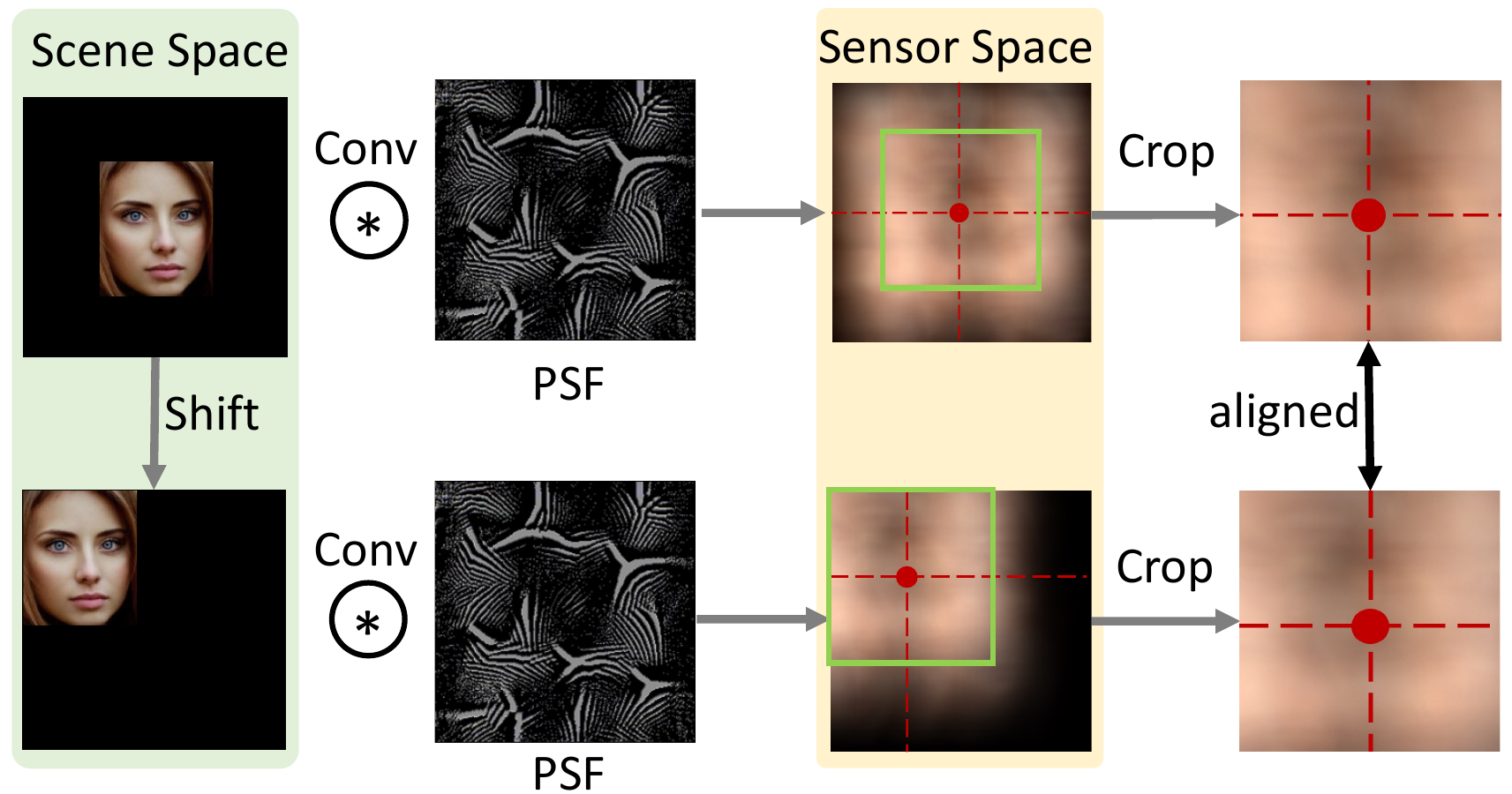}
    \caption{\textbf{Center alignment} for lensless capture of facial scenes.}
    \label{fig:alignment}
  \end{minipage}
  \hfill 
  \begin{minipage}[b]{0.47\linewidth}
    \includegraphics[width=\textwidth]{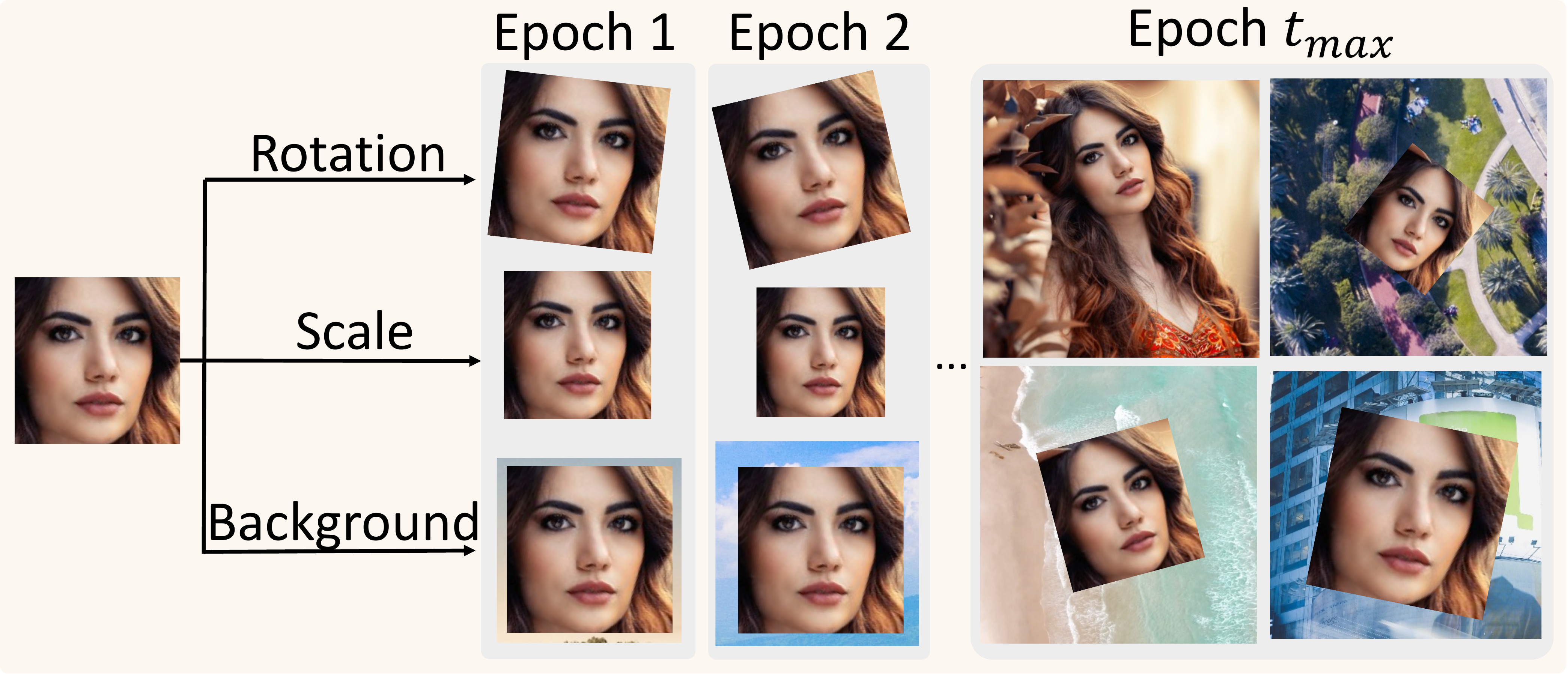}
    \caption{\textbf{Curriculum learning for augmentation}. The augmentation intensity gradually increases with training epochs.}
    \label{fig:CL}
  \end{minipage}
  \vspace{-2em}
  \label{fig:method_figure}
\end{figure}

\vspace{-1em}
\subsection{Alignment before Verification}
\vspace{-0.5em}

First, we explore the face-center alignment method, a preprocessing step for lensless captures before they are processed by the verification model. To start with, we define two spaces: \emph{scene space} and \emph{sensor space}, as shown in~\cref{fig:alignment}. Images before capture (optical mask) are in the \emph{scene space} and images captured by the sensor after light passes through the mask, are in the \emph{sensor space}. 

Unlike RGB images where the alignment is done by cropping faces from backgrounds, lensless images embed entire scenes across every pixel, making traditional alignment infeasible.
To solve this issue, we apply face-center alignment in the \emph{sensor space} to center and crop an initially off-centered encoded facial capture to a fixed size within the \emph{sensor space}. This alignment translates a face into a canonical position, reducing the training and testing gap and enhancing the accuracy of off-center faces.
In contrast to RGB-based face alignment which handles all types of similarity transformation, our lensless alignment only focuses on translation due to translation equivalence. Other similarity transformations, like rotation and scaling, are addressed in a later augmentation curriculum step.

The key idea of lensless face alignment is based on the principle of translation equivalence in convolution. According to this principle, the shift of the convolution output equals the convolution of the shifted input image, except on image boundaries. Recall that lensless capturing can be formulated as a convolution, as shown in \cref{eq:image_form}. Therefore, if we can determine the face center position in the \emph{scene space}, we can simply center the face in the \emph{sensor space}, as the centered sensor image (the bottom row of \cref{fig:alignment}) is the same as if we centered the face in the scene space and captured it (the top row of \cref{fig:alignment}).

Based on this, we train a face center detection model $\mathcal{G}_\phi$ on sensor images with the following objective, similar to~\cref{eq:target}:
\begin{equation}\label{eq:face-center}
    \min_{\phi,\theta_M}\sum_{i=1}^N\mathcal{L}(y_i,\mathcal{G}_\phi(g(I_i, M))),
\end{equation}
where $\{I_i,y_i\}_{i=1}^N$ is a face-center dataset with unaligned images $I_i$ and alignment $y_i$, and $M$ is the mask to be optimized. Then, we use the output of this detection model to center the face in sensor space.

\vspace{-1em}
\subsection{Verification Model Training}

After centering the face, we train a face verification model that can accurately recognize the face, under variations in rotation, scaling, and background changes. However, training a model to directly recognize encrypted captures under these conditions is difficult, often resulting in a model with poor generalization ability, as already verified by previous works on RGB-based face verification~\cite{taigman2014deepface}. To address this issue, we propose two strategies to make this training process tractable, the curriculum learning of augmentation and cross-modality distillation.

\textbf{Curriculum learning of augmentation.} To enhance the robustness under different variations, we first implement random augmentations, including rotation, scaling, and background changes, to the face images within the \emph{scene space}, as shown in \cref{fig:CL}. This enhances the diversity of our face recognition dataset. We then execute an end-to-end training process with the augmented data.

However, directly applying all augmentations to face images from scratch makes the training hard to converge. Therefore, we adopt an augmentation curriculum that progressively increases in difficulty. This enables the model to initially focus on acquiring face recognition capabilities through easy examples, and gradually develop robustness against variations with more challenging examples. We control the degree of augmentations, such as rotation angles and background sizes, and incrementally increase these augmentations, as illustrated in Fig.~\ref{fig:CL}. Specifically, we amplify the augmentation of variations following a cos-annealing warm-up strategy. Specifically, to augment a particular variation eventually with a maximum magnitude $\eta_{\max}$, we initiate the training with a minimum magnitude $\eta_{\min}$, and gradually increase the magnitude at each epoch $t$ as: 
\vspace{-0.5em}
\begin{equation}
\eta_t=\eta_{\min}+\frac{1}{2}\left(\eta_{\max}-\eta_{\min }\right)\left(1-\cos \left(\frac{t}{t_{max}} \pi\right)\right),
\vspace{-0.5em}
\end{equation}
where $t_{max}$ is the total number of epochs. Experiments in \cref{fig:result} show that this curriculum learning (CL) greatly improves the stability of the training process.

\textbf{Cross-modality distillation.}
Directly training face verification models on encoded images is very challenging, even when the impact of different variations is minimized through previous alignment and augmentation. 
This is because the encoded image does not keep the original spatial structure, making the training process hard to converge. Consequently, most previous methods reconstruct original images first before proceeding with downstream recognition tasks.

We propose a novel model distillation scheme to solve this challenge.
Model distillation is widely used in transferring knowledge from a well-trained \emph{teacher} model to another to-be-learned \emph{student} model\cite{hinton2015distilling}. In light of this, we propose a cross-modality distillation method to employ an already trained RGB-based face recognition model to distill knowledge into the hybrid optical-electric lensless model, enhancing its generalization capabilities and optimization efficiency. The RGB model extracts highly discriminative features for identity verification, and through distillation, we aim to empower our lensless model with similar discriminative capabilities. Considering the substantial difference in modality between lensless and RGB images, we apply relational knowledge distillation \cite{park2019relational} which focuses on transferring the structural relations of features, rather than the features themselves, to mitigate the effects of the modality gap. By transferring the relations of features, the inputs of teacher and student need not be pixel-matched as long as they represent the same concept. As shown in \cref{fig:pipeline}, the RGB \emph{teacher} model's output features of the aligned face are used to supervise the training of the hybrid optical-electric \emph{student} model with unaligned samples.

Moreover, distillation also brings additional advantages. It enhances the alignment of features from lensless captures at both instance-level and identity-level. The distillation process ensures that features from unaligned lensless images share a similar feature with the aligned RGB images and thus improves the robustness of the model under variations. Furthermore, given that RGB-based models proficiently associate different images of the same person with similar features, the distillation also makes features of lensless captures from a singular identity closely align with that identity's feature.

\textbf{Optimization objective.}
Based on our above discussion on training strategies, we define the optimization objective as the summation of the identification loss and the distillation loss. Specifically, given a face dataset $\{I_i,t_i\}_{i=1}^N$ with the identity labels $t_i$ and a pretrained RGB face recognition model $\mathcal{F}_t$, our training finds the optimal masks $M$ and the parameters $\theta$ for lensless face verification model, by minimizing the following loss:

\begin{equation}
       \min_{\theta,\theta_M}\sum_{i=1}^N \mathcal{L}_{id}(y_i,\mathcal{F}_\theta(\mathcal{C}(g(\mathcal{A}(t_i),M))))+\mathcal{L}_{dist}(\mathcal{F}_t(\hat{I}_i),\mathcal{F}_\theta(\mathcal{C}(g(\mathcal{A}(I_i),M)))).
\end{equation}
\vspace{-0.8em}

In the loss function, a RGB face image has sequentially gone through data augmentation $\mathcal{A}(\cdot)$, lensless imaging simulation $g(\cdot)$, and face alignment $\mathcal{C}(\cdot)$, resulting in the aligned capture of the augmented image $\mathcal{C}(\mathcal{S}(\mathcal{A}(I_i), M))$. The first term is the ArcFace identification loss~\cite{deng2019arcface} that compares the recognition result on this image with the ground truth $t_i$. The second term is a distillation loss that minimizes the difference in recognition results between the RGB and lensless pipelines, following the relational knowledge distillation framework~\cite{park2019relational}.

\vspace{-0.5em}
\section{Experiments}
In this section, we evaluate the performance and robustness of our end-to-end face verification pipeline, on both simulated and real-world data. Additionally, we analyze the contribution of each component, and also validate the robustness of the proposed approach under different lighting conditions. 

\begin{figure}[t]
   \vspace{-0.5em}
  \centering
  \begin{minipage}[b]{0.48\linewidth}
    \includegraphics[width=\textwidth]{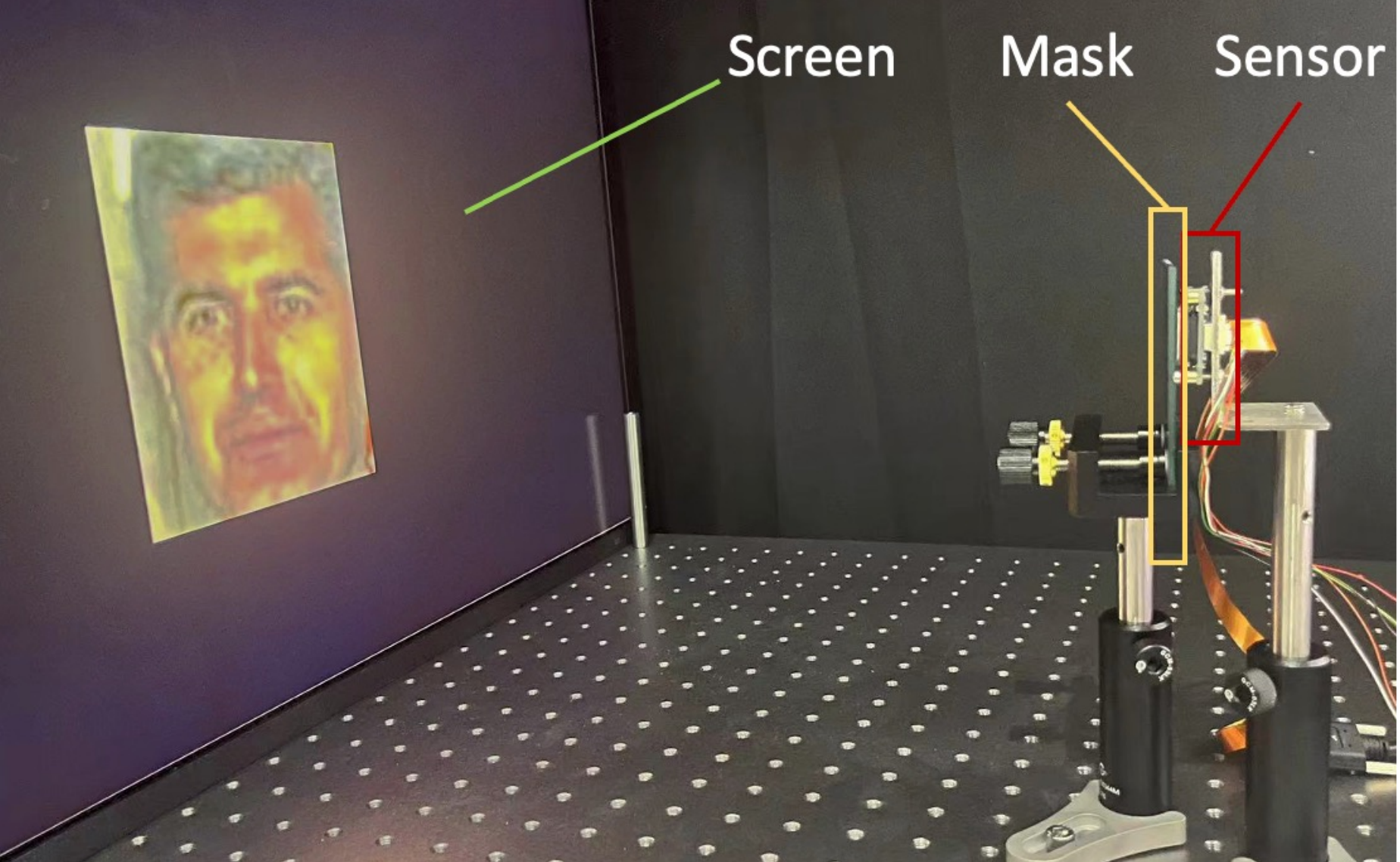}
    \vspace{-2.0em}
    \caption{\textbf{Prototype} of our lensless camera for face verification.}
    \label{fig:prototype}
  \end{minipage}
   \hfill 
  \begin{minipage}[b]{0.48\linewidth}
    \includegraphics[width=\textwidth]{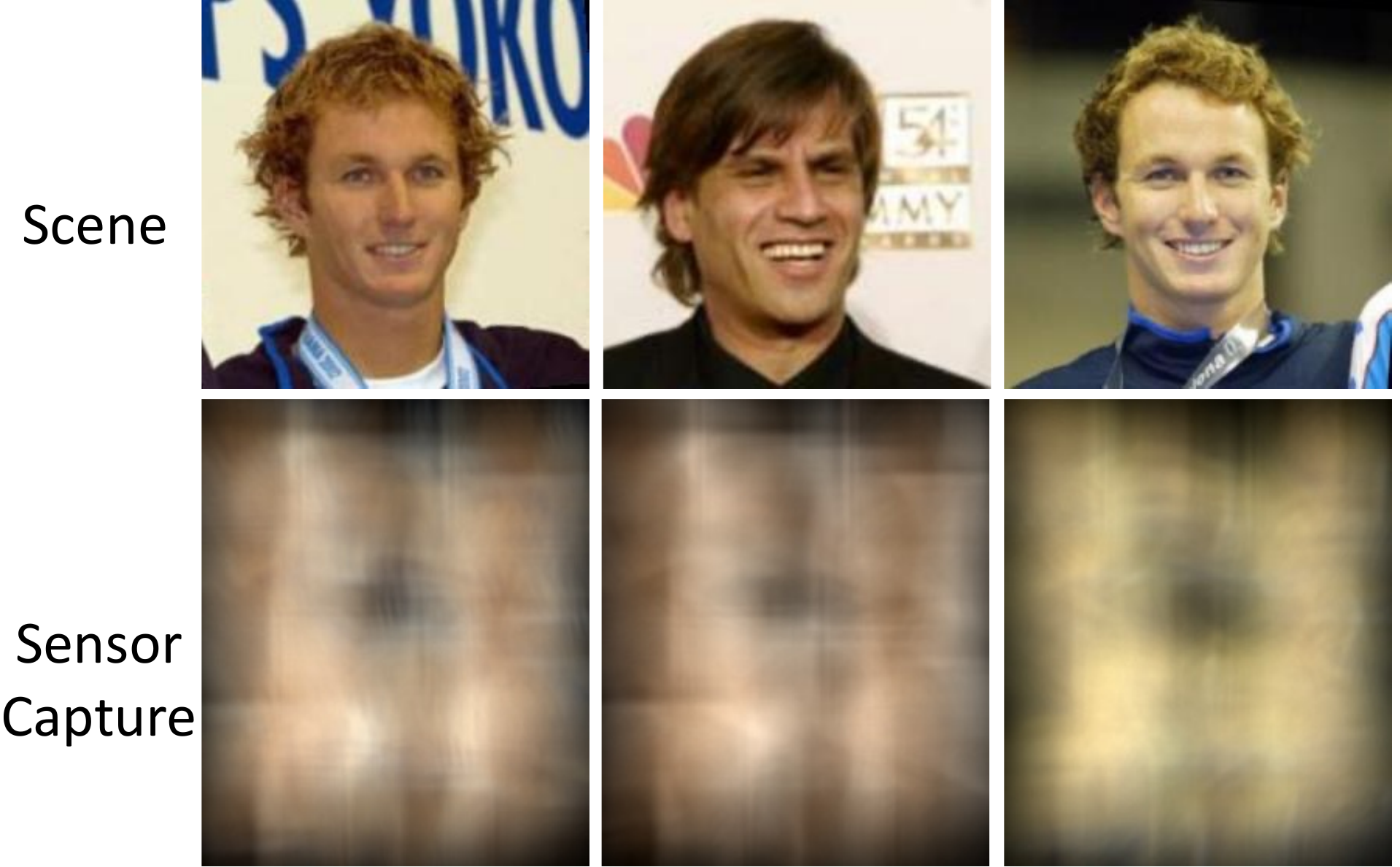}  
    \vspace{-2.0em}
    \caption{\textbf{Visualization} of scenes (top) and real captures (bottom).}
    \label{fig:privacy}
  \end{minipage}
  \vspace{-2em}
\end{figure}

\subsection{Experimental Setup}

\textbf{Hardware configuration.}
The experimental setup consisted of three main components: scenes displayed on a screen, a jointly-optimized binary amplitude mask, and a CMOS image sensor, as shown in Fig.~\ref{fig:prototype}. The image sensor used was a Sony IMX250, located inside a FLIR BFS-U3-51S5C-BD2 camera. It is a 2056$\times$2464 RGGB Bayer sensor, with each pixel being 3.45$\mu$m wide. The full sensor array measures approximately 7.1mm$\times$8.5mm.
The binary amplitude mask was fabricated on a chrome-coated glass substrate using photolithography.
The feature size (width of each unit grid) of the mask was 0.0276mm, around 8 times the pixel width. The mask has 257$\times$308 units, making its size approximately the same as the sensor. The mask-sensor distance is set to 2mm.

\textbf{Datasets.}
We utilized extensive public datasets for training and evaluating our proposed methods. 
For face verification tasks, we used the Asian-Celeb dataset \cite{facedata} containing 2.8M images from 94K identities for training and used both LFW~\cite{LFWTech} and FCFD~\cite{tan2018face} datasets for evaluation.  For both datasets, we followed the ``unrestricted with labeled outside data'' protocol proposed in \cite{huang2014labeled} and presented average accuracy rates on 10-fold cross-validation sets.

The optimization of the optical-electric model during the training phase was conducted entirely via simulation. However, our evaluation approach encompassed both simulated and real-world settings. In real-world evaluations, we capture the displayed facial scene on a screen with a lensless camera.

\textbf{Evaluation protocols.} 
We tried two evaluation protocols: aligned face evaluation and random face evaluation. In the aligned face evaluation, the algorithm takes the original aligned face in the dataset as input, retaining a constant face height of 27cm and a scene-to-camera distance of 50cm. For the random face evaluation, we further apply randomly perturbation to the facial image, including face center shifting of [-15cm, 15cm], face height scaling ranging from [22cm, 30cm], face rotating within angles [-30$^\circ$, 30$^\circ$], and scene-to-camera distance shifting between [40cm, 60cm]. Unlike the aligned face evaluation, the random face evaluation is factored into the scene background because the whole scene is larger than the face\footnote{The field of view of our camera is limited by the $2\theta_{CRA}$ (Chief Ray Angle) of the sensor \cite{boominathan2022recent}, which is 45$^\circ$ in our experiment. For example, if the scene-to-camera distance $d_{sc}$ = 50cm, the width of the whole scene is around $2d_{sc}\tan\theta_{CRA}$ $\approx$ 42cm.}. The background dataset BG-20K \cite{li2022bridging} is used to augment face background. To avoid data leakage, different background images are used for training and evaluation. Some augmentation examples are shown in Fig.~\ref{fig:CL}.

 \vspace{-1em}
\subsection{Implementation Details}
 \vspace{-.3em}

We adopted T2T-ViT \cite{yuan2021tokens} as the backbone of the face verification model. For efficient training, all lensless captures and optical convolution kernels were resized to 200$\times$240. To simulate the spatial intensity of scenes with RGB images, the face images were rescaled to specific sizes based on the physical settings (face height, scene-to-camera distance, mask-to-sensor distance). with details provided in the supplementary. We trained face verification models using the Adam \cite{kingma2014adam} optimizer with a learning rate $5 \times 10^{-4}$ until 50 epochs. The batch size is 64$\times$8. 

For face-center alignment, we followed ArcFace\cite{deng2019arcface} to normalize faces and defined the centers of normalized faces as the face centers. With the center, we center-cropped the lensless images to 60\% of their original size in both width and height. During training, we used the ground truth of the face center for alignment. For evaluation, we trained a separate face-center detection model to output the face-center positions. This model was trained with the optimized mask from the face verification model and the same backbone and training settings.

The augmentation curriculum was implemented during the initial 20 epochs of the 50-epoch training period. The training began with augmentation parameters (shift, rotation, scale, background) identical to those in the aligned face evaluation setting. Over these 20 epochs, we progressively adjusted these parameters to align with those employed in the random face evaluation setting.

To implement cross-modality distillation, we first trained an RGB teacher model with aligned RGB images from the training dataset. Subsequently, we trained the optoelectronic (lensless) model on the same dataset, this time with lensless optics simulation, and incorporated distillation loss to enhance the learning process. 
It's noted that both the RGB teacher model and the lensless student model share the same model architecture, except for the simulated optics.

\begin{figure}[htbp]
   \vspace{-0.5em}
  \centering
  \begin{minipage}[b]{0.46\linewidth}
  \captionof{table}{Decoded Image quality for the varying number of plaintext attacks.} \label{tab:decoded_iq}
        {\fontsize{7.2pt}{\baselineskip}\selectfont \fontfamily{ptm}\selectfont
        \begin{tabular}{c|cccc}
        \hline
       
        \#attacks & 10 & 100 & 1,000   & 10,000 \\\hline
        PSNR/SSIM & 12.1/0.31  &  13.9/0.39      &   16.2/0.50 &   19.1/0.59     \\ \hline
        \end{tabular}
        }
    \includegraphics[width=\textwidth]{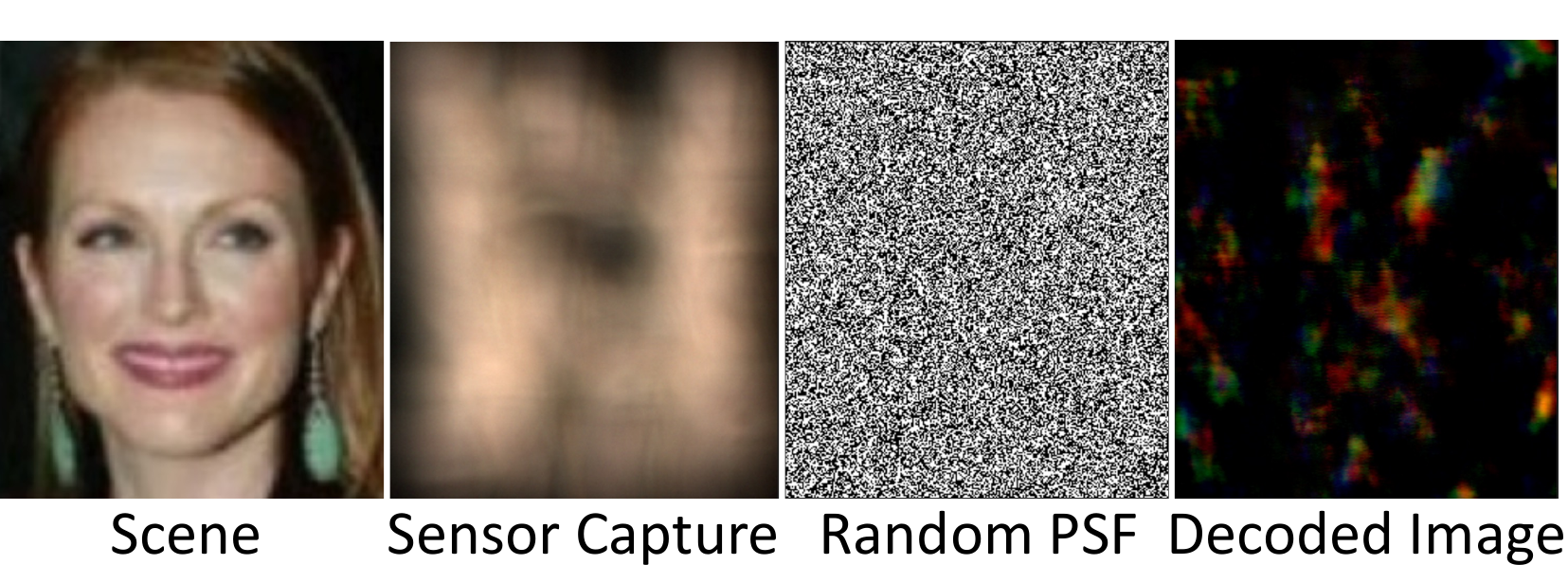}
     \vspace{-1em}
    \captionof{figure}{Example of decoding lensless captures using a random PSF.}
    \label{fig:random_decode}
  \end{minipage}
  \hfill 
  \begin{minipage}[b]{0.46\linewidth}
  \centering
    \includegraphics[width=0.9\textwidth]{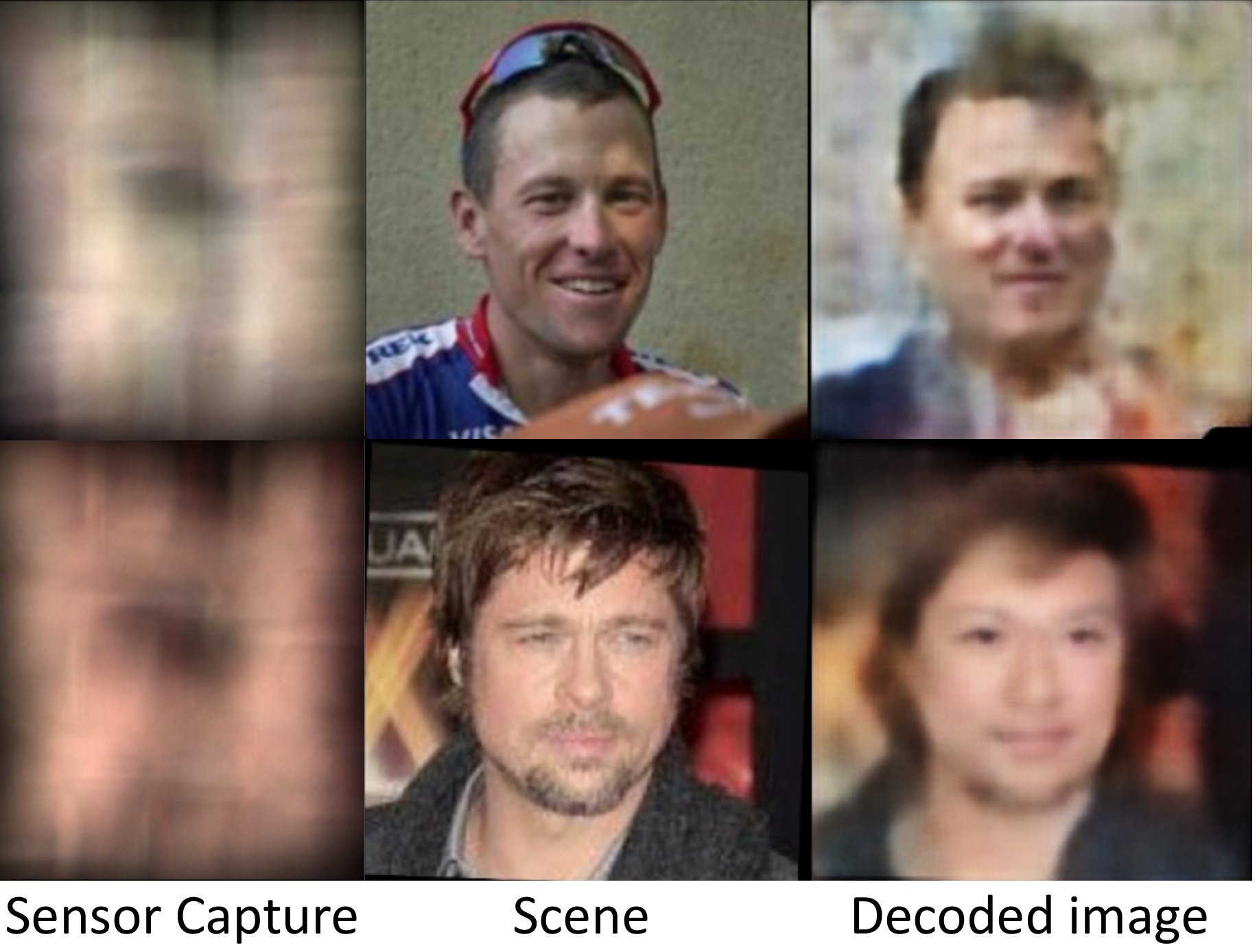}
    \vspace{-1em}
    \caption{Decoded examples by decoder trained with 10,000 attack pairs.}
    \label{fig:attack_decoded_example}
  \end{minipage}
  \vspace{-2em}
\end{figure}

\vspace{-1em}
\subsection{Privacy Protection with Learned Mask}
\vspace{-.5em}

To demonstrate the proposed does protect user privacy, we did the following experiments. First, we showed that the raw lensless captures are unrecognizable to both humans and normal RGB-based face verification models. Fig.~\ref{fig:privacy} presents some visual examples from our lensless camera. The lensless captures of two faces with similar backgrounds appear highly similar to each other (left and middle ones), while captures of the same face under different backgrounds appear quite distinct (left and right ones).
This observation indicates that the raw lensless data lack distinguishable cues for human perception of identity, as variations in background overwhelm subtle differences between faces. Quantitatively, a conventional RGB-based face verification model only achieves 51.2\% accuracy (50\% by random guessing) on lensless captures from the LFW dataset, further validating its potential for preserving privacy.

Moreover, we demonstrate that our lensless imaging system also preserves privacy against reconstruction attacks. We assumed attackers cannot access the camera's point spread function (PSF)\cite{privacy2020workshop}, and evaluated two attack scenarios:
i)\textbf{Attackers only access lensless captures only:} In this setup, we utilized 50 random mask patterns as surrogate PSFs and solved the inverse imaging problems with ADMM\cite{boyd2011distributed}. The reconstructed images yielded very low average quality, with only 11.7dB PSNR and 0.25 SSIM. Some reconstruction examples are shown in \cref{fig:random_decode}. These results convincingly indicate that random PSF guesses fail to reconstruct the origin scenes meaningfully.
ii)\textbf{Attackers access to RGB-lensless face pairs}: In this setup, we utilized RGB-lensless pairs, created from the LFW dataset, to train a U-Net \cite{ronneberger2015u} as a decoder for simulating reconstruction attacks. The performance of the trained decoder with the varying numbers of pairs, as measured by image quality (PSNR/SSIM) on the left LFW dataset, is summarized in \cref{tab:decoded_iq}. The results indicate that with fewer than 1,000 pairs, the quality of the reconstructed images is very low, showing that attackers can hardly recover the original scene with a limited number of plaintext attacks. We also show some visual examples of reconstruction results by the decoder trained with 10,000 pairs in \cref{fig:attack_decoded_example}. The visual results show that humans can hardly recognize facial identity, further demonstrating that privacy protection ability.

\vspace{-0.3em}
\subsection{Comparison with Face Verification Methods}
To evaluate the face verification performance with the lensless camera under both constrained and unconstraint conditions, we compare our methods with other state-of-the-art face verification methods under two protocols: aligned face evaluation (A)  and random face evaluation (R).

\begin{table}[t]
\caption{
\textbf{Comparison with lensless face verification methods}. 
Input type refers to the input data type for face verification models. 
}
\vspace{-1em}
\label{tab:comparison}
\setlength\tabcolsep{2.8pt}
\scriptsize
\centering
\begin{threeparttable}
\begin{tabular}{cc|c|c>{\columncolor{gray!20}}c|ccc>{\columncolor{gray!20}}c}
\toprule
\multicolumn{2}{c|}{Methods} & WebCam\dag & FlatCam & Ours & FlatCam & FlatVer & FlatDCT & Ours \\
\midrule
\multicolumn{2}{c|}{Input type} & \makecell[c]{Captured\\RGB} & \makecell[c]{Reconstructed\\RGB (real)*} & \makecell[c]{Lensless\\(real)} & \makecell[c]{Reconstructed\\RGB (sim.)} & \makecell[c]{Lensless\\(sim.)} & \makecell[c]{Lensless\\(sim.)} & \makecell[c]{Lensless\\(sim.)} \\
\midrule
\multirow{4}{*}{{\rotatebox{90}{Datasets}}} 
& LFW (A) & 99.32 & - & 91.51 & 92.18 & 68.13 & 80.13 & \textbf{96.78} \\
& LFW (R) & 98.82 & - & 89.88 & 90.91 & 63.80 & 66.25 & \textbf{94.98} \\
\cmidrule{2-9}
& FCFD (A) & 98.31 & 82.15 & 90.23 & 87.33 & 57.35 & 77.81 & \textbf{94.34} \\
& FCFD (R) & 97.90 & 80.75 & 87.45 & 85.98 & 51.05 & 60.87 & \textbf{92.81} \\
\bottomrule
\end{tabular}
\begin{tablenotes}
\scriptsize
   \item \dag Upper Bound. \quad * We use the published real-captured dataset~\cite{tan2018face} for evaluation. 
\end{tablenotes}
\end{threeparttable}
\vspace{-2em}
\end{table}

Lensless face verification methods can be divided into two classes: 2-step reconstruct-and-verify and 1-step verify from the lensless capture. (1) For the 2-step method, \emph{WebCam} is a traditional RGB-based face verification method. \emph{FlatCam} \cite{tan2018face} reconstructs the face images from encoded images captured by flatcam \cite{asif2016flatcam}, one type of lensless camera, and then verifies faces on reconstructed images. For 2-step random face evaluation, face detection~\cite{qi2022yolo5face} and alignment~\cite{bulat2017far} are conducted before verification.
(2) For the 1-step method, \emph{FlatVer} is a network that directly verifies faces on raw flatcam data without reconstruction. \emph{FlatDCT} \cite{henry2023privacy} is a method that transforms the flatcam captures into a multi-resolution DCT subband representation and then verifies faces on this DCT representation. The training of 1-step networks uses a similar augmentation as in our random face evaluation. For a fair comparison, we reproduce all these methods using the same backbone as our methods and we use the same training dataset to train the RGB teacher in our distillation method, without using extra training data.

Results in~\cref{tab:comparison} show that our method outperforms both the 2-step and 1-step methods by a significant margin, on both synthetic and real data. The performance gap between the simulated and the real experiment is caused by various factors, including inconsistencies in the image processing pipeline, noise in ambient lighting, inaccuracies in the physical simulation model, and errors in the hardware setup parameters. This performance gap is common in lensless experiments\cite{khan2020flatnet,shi2022loen}, and our gap is in an acceptable region. Even with this performance gap, our model on real images outperforms \emph{FlatVer} and \emph{FlatDCT} on synthetic ones and achieves better performance with \emph{FlatCam} on real images.
All these results validate the effectiveness of our method.
\vspace{-1em}
\subsection{Ablation Study}
\vspace{-.5em}
We investigate the effectiveness of each component in our method under two evaluation settings: aligned face evaluation and random face verification.

\begin{table}[t]
  \centering
\caption{\textbf{Ablation studies} of (a) aligned and (b) random face evaluation.}
\vspace{-1em}
\begin{minipage}{0.36\textwidth}
\setlength\tabcolsep{2.5pt}
\label{tab:ablation}
\renewcommand{\arraystretch}{1.1}
\scriptsize
\centering
(a) Aligned face evaluation.
\begin{tabular}{cc|cc}
\toprule
\multicolumn{2}{c|}{Methods}  & \multicolumn{2}{c}{Datasets} \\ \hline
\makecell[c]{Mask \\ optim.\\} & \makecell[c]{Cross-modal \\ distillation\\} & \makecell[c]{LFW\\(A)} & \makecell[c]{FCFD\\(A)} \\ \midrule
   $\times$       & $\times$  &   68.77    &    63.15   \\
$\times$    &  $\checkmark$       & 92.70  &  88.18  \\
$\checkmark$    &  $\times$        & 71.33   & 64.82  \\
   $\checkmark$    &  $\checkmark$     &  \textbf{96.78}     &  \textbf{94.34}     \\
\bottomrule
\end{tabular}
\end{minipage}
\hfill
\begin{minipage}{0.61\textwidth}
\setlength\tabcolsep{0.8pt}
\scriptsize
\centering
(b) Random face evaluation.
\begin{tabular}{cccc|cc}
\toprule
\multicolumn{4}{c|}{Methods}  & \multicolumn{2}{c}{Datasets} \\ \hline
\makecell[c]{Mask \\ optim.\\} & \makecell[c]{Face-center \\ alignment\\}   & \makecell[c]{Augmentation \\ curriculum\\}   &  \makecell[c]{Cross-modal \\ distillation\\} & \makecell[c]{LFW\\(R)} & \makecell[c]{FCFD\\(R)} \\ 
\hline
 $\times$  &  $\checkmark$   &     $\checkmark$    &   $\checkmark$    &  91.36     &   82.43    \\
$\checkmark$    &  $\times$   &     $\checkmark$    &   $\checkmark$    &  84.77   &   79.95    \\
$\checkmark$    &  $\checkmark$  &      $\times$    &   $\checkmark$       &    83.65     &    80.32   \\ 
$\checkmark$    &  $\checkmark$  &    $\checkmark$      &       $\times$   & 73.45      &   59.97    \\
$\checkmark$ &   $\checkmark$    &     $\checkmark$    &  $\checkmark$     &   \textbf{94.98}    &  \textbf{92.81}     \\
\bottomrule
\end{tabular}
\end{minipage}
\vspace{-1em}
\end{table}

\begin{figure}[t]
\centering
\begin{minipage}{0.36\textwidth}
\centering
\includegraphics[width=0.75\linewidth]{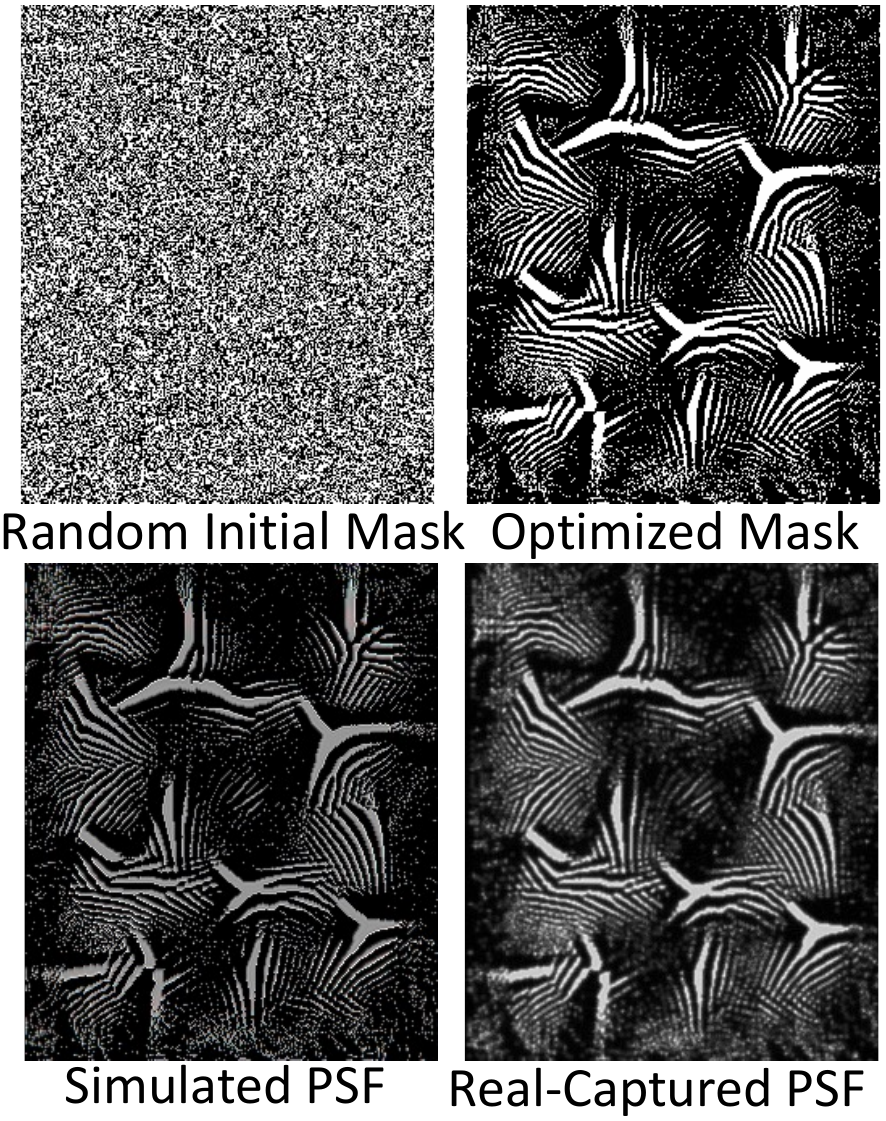}
\vspace{-1em}
\caption{Top: Mask shape before and post optimization. Bottom: Simulated PSF is close to the real-captured PSF.}
\label{fig:psf}
\end{minipage}
\hfill
\begin{minipage}{0.6\textwidth}
\includegraphics[width=\linewidth]{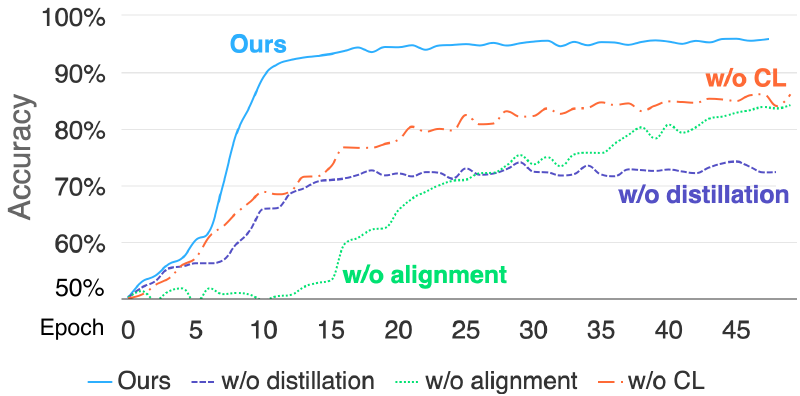}
\vspace{-1em}
\caption{Comparison of test accuracy of our method against ablated variations (CL refers to curriculum learning) on LFW (R) dataset at different training epochs.}
\label{fig:result}
\end{minipage}
\vspace{-.5em}
\end{figure}

In the aligned face evaluation, we first show how mask optimization generates better masks compared with randomly initialized baselines, and then demonstrate the effectiveness of distillation. \cref{fig:psf} illustrates how the mask optimization changes its shape and corresponding PSF. ~\cref{tab:ablation}(a) shows that mask optimization brings 3-4\% performance boosting and the the proposed distillation significantly boosts the accuracy by about 25\%.

In the random face evaluation, we ablate all four key components: 1) mask optimization, 2) face-center alignment, 3) augmentation curriculum, and 4) cross-modality distillation. \cref{tab:ablation}(b) shows that all four components have positive contributions, with the distillation being the most important one. Interestingly, even with a random unlearned mask, the other components still provide reasonable face verification results.
We also visualize the epoch-validation accuracy curves of our procedures in \cref{fig:result}, when individual components are omitted.

This experiment demonstrated that naively training for our end-to-end pipeline may result in poor performance, as we claimed in the introduction. All four key components are important to the robustness of our system. Without any one part, the accuracy drops below 85\%, making the whole system infeasible.
\begin{figure}[htbp]
   \vspace{-0.5em}
  \centering
  \begin{subfigure}[b]{0.62\linewidth}
  \centering
    \includegraphics[width=0.9\textwidth]{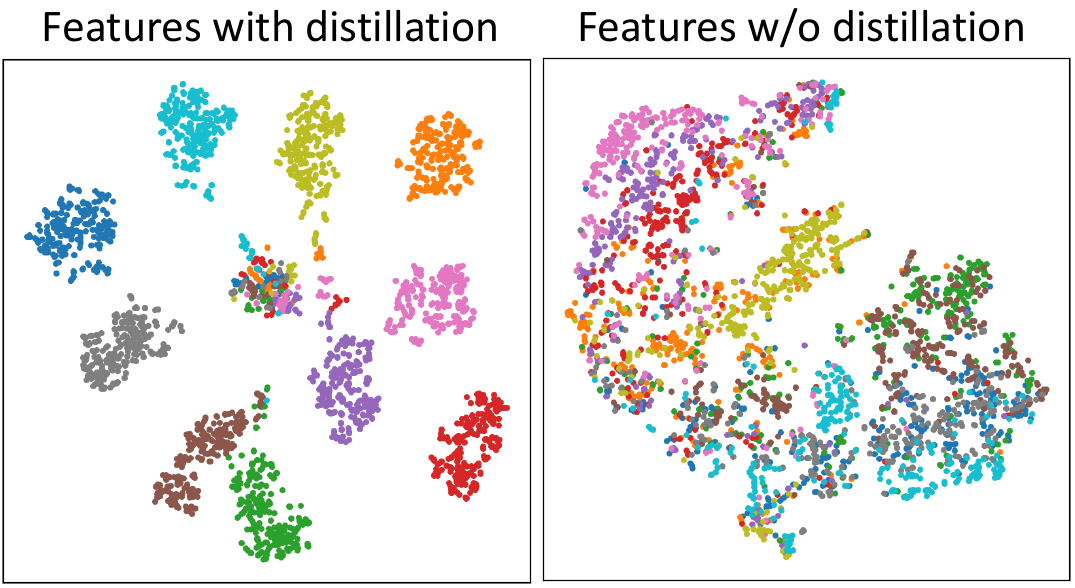}
    \caption{Visualization for face features. Different colors represent different identities.}
    \label{fig:tsne}
  \end{subfigure}
  \begin{subfigure}[b]{0.37\linewidth}
    \includegraphics[width=0.9\textwidth]{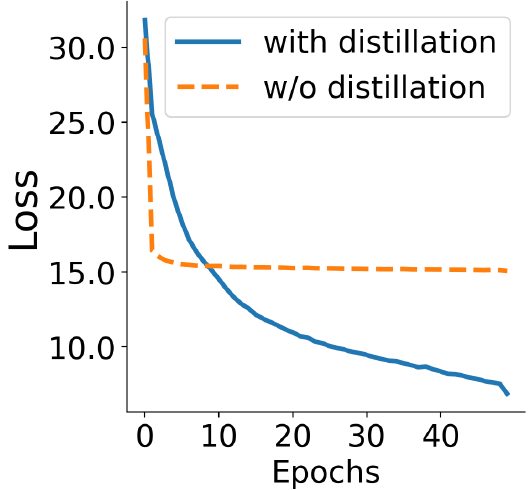}  
    \caption{Loss over epochs.}
    \label{fig:loss}
  \end{subfigure}
  \vspace{-2em}
  \caption{\textbf{Ablation study} of distillation. (a) Features for different identities are better grouped after distillation. (b) Distillation improves the convergence.}
  \vspace{-2em}
  \label{fig:distill}
\end{figure}

\vspace{-1em}
\subsection{Discussion about Cross-modality Distillation} 
Our ablation study indicates that cross-modality distillation is a critical component for optimization. To demonstrate this, we visualize the facial features of ten randomly selected individuals from the FCFD (R) dataset using t-SNE \cite{van2008visualizing} for dimensionality reduction. We compare these visualizations between models with and without cross-modality distillation. \cref{fig:tsne} shows that cross-modality distillation effectively separates features from different identities, enhancing the performance of face verification tasks.
Furthermore, we plot the face verification loss curves for both models during training, as depicted in \cref{fig:loss}. The comparison reveals that the model lacking distillation tends to converge to a local minimum with poor generalization capabilities. In contrast, incorporating distillation allows the model to avoid these local minima during optimization.

Our distillation, coupled with the novel augmentation and alignment, enables our model to learn transformation-invariant features, evidenced by consistent performance in both aligned and random evaluations in \cref{tab:comparison}. In contrast, relying on distillation and normal augmentation alone, alternative approaches such as FlatVer and FlatDCT \cite{henry2023privacy} show limited accuracy, achieving only 75.6\% and 78.4\% on LFW(R) and 66.0\%  and 72.6\% on FCFD(R) respectively.

\vspace{-1em}
\subsection{Robustness to Varying Lighting Conditions}
\vspace{-.5em}

Our lensless face verification system can handle diverse lighting conditions because the training simulation incorporates various noise levels. Specifically, we introduce noise with an average signal-to-noise ratio (SNR) of 30dB during training using the camera noise model~\cite{hasinoff2014photon}. On the LFW(A) dataset, the verification accuracy under SNR 20dB, 25dB, 30dB, 35dB, and 40dB are 94.20\%, 96.03\%, 96.78\%, 96.62\%, and 96.68\%, respectively. This result shows the robustness of our model against different lighting conditions. Furthermore, optimizing the electronic part of our lensless model for low-light conditions (SNR=20dB) can further improve the accuracy (from 94.20\% to 96.02\%). This optimization further demonstrates the adaptability and efficiency of our approach in challenging lighting scenarios. More details can be found in the supplementary.

\vspace{-1em}
\section{Conclusion and Limitation}
\vspace{-0.7em}
We introduce an innovative end-to-end optimization approach for robust face verification that operates directly on raw lensless captures. This method enhances privacy, accuracy, and processing speed, representing a significant advancement over traditional two-stage lensless face verification. We validate our approach through simulations and further confirm its effectiveness with a real-world lensless camera prototype. When compared with traditional RGB-based face verifications, the lensless face verification system provides inherent hardware-level privacy protections with a small, efficient, and low-cost device and presents an attractive performance tradeoff. One limitation is that there is still a performance gap between simulations and real-world environments, as one of the common problems of lensless networks trained on simulated data.


%
%
\bibliographystyle{splncs04}
\bibliography{main}
\end{document}